%% file: paper.tex
\PassOptionsToPackage{table}{xcolor}
\documentclass{fairmeta}
\usepackage{amsmath,amsfonts,amssymb}
\usepackage{algorithmic}
\usepackage{algorithm}
\usepackage{array}
\usepackage{textcomp}
\usepackage{url}
\usepackage{verbatim}
\usepackage{threeparttable}
\usepackage{amsthm}
\newtheorem{theorem}{Theorem}
\usepackage{enumitem}
\usepackage{pifont}
\usepackage{xspace}
\input{math_commands.tex}

\def\eqref#1{(\ref{#1})}
\newcommand{\method}{SalQ-VLM\xspace}

\newcommand{\co}{C_o}
\newcommand{\ci}{C_i}
\definecolor{bgblue}{RGB}{99,202,225}
\definecolor{myblue}{RGB}{16,56,115}
\definecolor{myred}{RGB}{115,1,0}
\definecolor{myyellow}{RGB}{168,131,55}
\definecolor{mygreen}{RGB}{7,44,2}
\definecolor{mygray}{RGB}{128,127,127}

\title{SalQ-VLM: Fine-Grained Saliency-Guided Quantization for Vision-Language Models}
\author[1]{Yufei Xue}
\author[1]{Yushi Huang}
\author[1]{Lunjie Zhu}
\author[2,3]{Jiawei Shao}
\author[1]{Jun Zhang}
\affiliation[1]{The Hong Kong University of Science and Technology, Hong Kong}
\affiliation[2]{School of Artificial Intelligence, OPtics and ElectroNics (iOPEN), Northwestern Polytechnical University}
\affiliation[3]{Institute of Artificial Intelligence (TeleAI), China Telecom, China}
\metadata[Correspondence to]{Jun Zhang (\email{eejzhang@ust.hk}); Jiawei Shao (\email{jiawei.shao@nwpu.edu.cn})}
\metadata[Keywords]{Vision-language models, model compression, post-training quantization}

\begin{document}

\abstract{
Large language models (LLMs) have demonstrated remarkable capabilities across diverse language tasks, motivating their extension to vision-language models (VLMs) for multimodal understanding. However, billion-parameter VLMs incur substantial memory and computational costs that hinder deployment in resource-constrained settings. Post-training quantization (PTQ) compresses models and accelerates inference without retraining, yet remains underexplored for VLMs. We identify two intrinsic VLM activation properties in PTQ: (1) visual over-representation, where vision tokens are excessive and often redundant, and (2) the modality gap separating text and vision tokens in the latent feature space. Prior methods largely overlook these properties, leading to quantization performance degradation. To address this mismatch, we propose \method, an importance-aware PTQ framework that prioritizes salient tokens and suppresses redundant vision tokens during calibration. We derive a gradient-driven importance factor that captures token-level importance variance and is theoretically grounded in the relationship among loss perturbation, activation errors, and output gradients. \method obtains this factor through a single lightweight block-wise gradient-caching pass and incorporates it into the layer-wise reconstruction objective. Because \method modifies only calibration, it adds no inference-time operations and remains compatible with existing high-performance kernels. Extensive evaluations across benchmarks and backbones show that \method consistently outperforms strong PTQ baselines, especially under ultra-low-bit quantization. Notably, it improves MME-RealWorld accuracy by 16.45\% under INT2g128 quantization.
}

\maketitle

\section{Introduction}\label{sec:introduction}
Large language models (LLMs) \citep{bubeck2023gpt4,touvron2023llama,touvron2023llama2,deepseek2025deepseek-r1,jiang2023mistral} have achieved remarkable progress across diverse natural language processing tasks, which has motivated the development of vision-language models (VLMs) \citep{wang2024qwen2vl,bai2025qwen25vl,li2024llava-ov,chen2024internvl2,xiaomi2025mimovl,zhu2025internvl3} that process multi-modal inputs, including text, images, and videos. Despite their strong capabilities, rapid model scaling complicates their deployment in resource-limited settings \citep{an2026aiflow,shao2025aiflow,yuan2025task}.

\begin{figure}[t]
    \centering
    \includegraphics[width=0.6\linewidth]{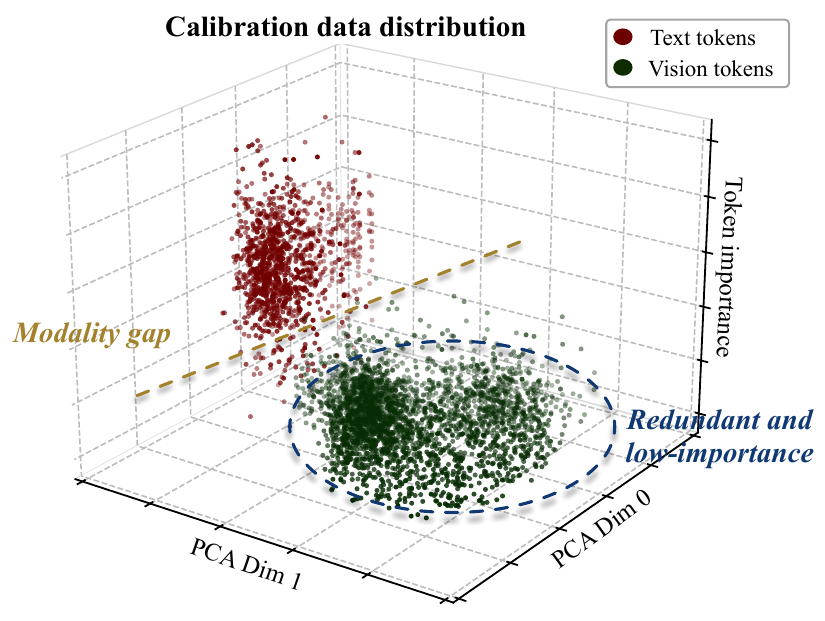}
    \caption{Token-level importance and PCA-distribution~\citep{shlens2014pca}. The plotted tokens are activations from the query projection of the 20-th layer in Qwen2-VL-7B-Instruct.}
    \label{fig:teaser}
\end{figure}

Quantization offers an effective remedy by converting full-precision weights and activations (\emph{e.g.} FP16/BF16) into reduced-precision formats (\emph{e.g.} INT8/INT4), thereby reducing memory footprint and computational cost. For example, serving a 70B-parameter model in FP16 requires at least 140 GB of GPU memory. INT4 quantization reduces this footprint sufficiently to fit the model on a single 40 GB A100 GPU~\citep{lin2023awq}. Post-training quantization (PTQ) has become a prevalent deployment strategy because it incurs minimal computational overhead and avoids costly fine-tuning or retraining. For LLMs, existing PTQ methods are commonly grouped into three families. Smoothing-based approaches apply equivalent transformations to suppress activation outliers~\citep{lin2023awq,shao2023omniquant}. Representative Hessian-informed methods such as GPTQ~\citep{frantar2022gptq} minimize layer-wise quantization error through theoretically grounded weight compensation~\citep{frantar2022gptq,li2025gptqv2,guan2024aptq}. Rotation-based methods further reshape weight distributions into quantization-friendly forms via Hadamard transforms~\citep{ashkboos2024quarot,liu2024spinquant,hu2025ostquant}. Despite their effectiveness, these approaches often require dedicated kernel support and framework adaptation. Although substantial progress has been made for LLMs, PTQ for VLMs remains underexplored. More recently, MBQ~\citep{li2025mbq}, MQuant~\citep{yu2025mquant}, and QSLAW~\citep{xie2024qslaw} have introduced modality-aware strategies for VLM quantization. QSLAW incorporates quantization-aware scale learning into vision-language instruction tuning. Despite its effectiveness, it is resource-intensive and requires substantial data-engineering effort. MQuant enables full-static quantization through Hadamard rotations, but its deployment relies on specialized kernel implementations. MBQ introduces a modality-balanced optimization objective but remains limited by a suboptimal grid-search process. Overall, existing methods fail to provide an effective PTQ solution for both calibration and inference.

In this paper, we pioneer the characterization of visual over-representation and the modality gap as two central challenges in VLM PTQ. As shown in Figure~\ref{fig:teaser}, VLM activations exhibit pronounced visual over-representation and a clear modality gap between text and vision tokens in the latent feature space. Specifically, multi-modal calibration data contain far more vision tokens than text tokens, and many vision tokens are redundant rather than informative. These two properties have been largely overlooked by prior PTQ methods. Existing LLM PTQ approaches minimize the layer-wise reconstruction loss while treating all tokens uniformly~\citep{lin2023awq,frantar2022gptq,li2025gptqv2}. This token-agnostic design therefore biases quantization toward dominant but redundant visual features. Although recent VLM PTQ methods incorporate modality information~\citep{li2025mbq,yu2025mquant,xie2024qslaw}, they still lack fine-grained token-level importance estimation and thus cannot fully counteract redundancy-induced calibration bias.

To address this mismatch, we propose \method, an importance-aware PTQ framework tailored to VLMs. \method introduces a gradient-driven importance factor $\rmG$ that assigns higher weights to salient tokens and suppresses redundant vision tokens during calibration. We establish a theoretical connection between loss perturbation and token-level activation error, which justifies using gradient statistics to estimate token importance. To make this estimation practical, \method performs a single lightweight block-wise backpropagation at each decoding block and derives layer-specific importance factors without parameter updates. These factors are then incorporated into an importance-aware layer-wise calibration objective that can be seamlessly integrated into mainstream PTQ pipelines. As a result, \method remains training-free and introduces no additional inference overhead. The core contributions are summarized as follows:
\begin{itemize}[leftmargin=*, noitemsep, label=$\triangleright$]
    \item We pioneer the characterization of visual over-representation and the modality gap as two overlooked PTQ challenges in VLMs, and uncover a fundamental mismatch between vision redundancy and token-agnostic calibration objectives. We further verify that this neglected redundancy disrupts layer reconstruction and explains the degraded performance of directly applying LLM PTQ methods to VLMs.
    \item We propose a gradient-driven importance factor $\rmG$ to suppress less informative tokens during calibration. Its effectiveness is both \textit{theoretically} and \textit{experimentally} supported through the established link between loss perturbation and first-order token-level activation errors. Moreover, its efficiency is ensured via a lightweight \textit{block-wise} backpropagation scheme.
    \item Through extensive experiments, we validate the superior performance of our framework on VLMs, particularly under ultra-low-bit quantization regimes. Notably, \method significantly suppresses the baselines and yields accuracy improvements of up to 16.45\% under 2-bit quantization.
\end{itemize}

The remainder of this paper is organized as follows. Section~\ref{sec:related} reviews VLMs and existing PTQ methods. Section~\ref{sec:preliminary} introduces the necessary background and presents our motivation. Section~\ref{sec:method} details \method: the gradient-driven importance factor, efficient block-wise gradient caching, and importance-aware calibration. Section~\ref{sec:experiments} reports the main results, ablation studies, and efficiency evaluation. Finally, Section~\ref{sec:conclusion} concludes the paper and discusses its limitations.

\section{Related Works}\label{sec:related}
In this section, we discuss related works related to our paper.
We first review the VLMs and then review the existing PTQ methods for LLMs and VLMs.

\subsection{Vision-Language Models} 
VLMs extend LLMs with visual encoders and cross-modal projection modules, enabling unified reasoning over text, images, and videos~\citep{li2024llava-ov,wang2024qwen2vl,bai2025qwen25vl,xiaomi2025mimovl,chen2024internvl2,zhu2025internvl3}. Early VLMs primarily focused on single-image understanding, whereas recent models increasingly support multi-image inputs, videos, dynamic resolutions, and long multimodal contexts. LLaVA-OneVision~\citep{li2024llava-ov}, for example, unifies single-image, multi-image, and video tasks within one model and demonstrates strong transfer across these scenarios.

Recent VLM families further improve visual flexibility and model scalability. Qwen2-VL~\citep{wang2024qwen2vl} introduces naive dynamic resolution, which maps inputs of different resolutions to variable numbers of vision tokens. Its multimodal rotary position embedding jointly models positional information from text, images, and videos. Qwen2.5-VL~\citep{bai2025qwen25vl} extends this design with a native dynamic-resolution vision transformer, window attention, and absolute temporal encoding. These components strengthen document parsing and long-video understanding. The InternVL series explores complementary scaling strategies across model size, data, and test-time computation~\citep{chen2024internvl2}. InternVL3~\citep{zhu2025internvl3} further adopts native multimodal pretraining and variable visual position encoding to support extended contexts. Although these advances substantially improve multimodal perception and reasoning, they also produce longer and more variable visual token sequences while relying on increasingly large language backbones. Consequently, efficient compression and deployment have become central challenges for modern VLMs.

\subsection{Post-Training Quantization} 
PTQ has become a widely adopted technique for compressing~\citep{li2024tesseraq, wnag2024ptsbenchcomprehensiveposttrainingsparsity, huang2025qvgenpushinglimitquantized, gong2024llmcbenchmarkinglargelanguage} and accelerating large models~\citep{lv2024ptq4sam, chen2024db-llm, tian2024qvd, huang2025harmonicaharmonizingtraininginference}. In the context of LLMs, approaches such as AWQ~\citep{lin2023awq} and SmoothQuant~\citep{xiao2023smoothquant} address activation outliers through smoothing-based transformations. Hessian-informed methods~\citep{frantar2022optimal-brain-compression,frantar2022gptq,li2025gptqv2} provide closed-form layer-wise quantization schemes. More recently, rotation-based methods~\citep{ashkboos2024quarot,liu2024spinquant,hu2025ostquant} reshape parameter distributions to improve low-bit quantization robustness. However, PTQ becomes more challenging for VLMs because text and vision tokens exhibit distinct statistical properties, and VLM-specific solutions remain limited. MQuant~\citep{yu2025mquant} proposes modality-specific static quantization and attention-invariant flexible switching to address the modality gap and visual outliers. MBQ~\citep{li2025mbq} observes error-sensitivity variance across modalities and improves AWQ~\citep{lin2023awq} through scale search. QSLAW~\citep{xie2024qslaw} learns quantization step sizes with modality-aware warmup to alleviate overfitting. Unlike these modality-oriented methods, Q-VLM~\citep{wang2024q-vlm} captures cross-layer dependency in VLMs and partitions blocks based on entropy.

Despite these advances, existing methods still leave a clear gap for practical VLM quantization. LLM-oriented PTQ approaches are largely designed for homogeneous text tokens and therefore transfer poorly to multimodal activation patterns. Moreover, few VLM PTQ methods explicitly account for visual over-representation and the modality gap between text and vision tokens. As a result, an efficient and effective VLM PTQ solution that supports accurate calibration and low-overhead inference remains open.

\subsection{VLM Inference Acceleration}
Beyond quantization, VLM inference can be accelerated by reducing the computation associated with long visual sequences. A major direction is training-free visual token compression. Importance-based methods identify and remove tokens that contribute less to subsequent reasoning. FastV~\citep{chen2024fastv} uses language-model attention to prune low-importance vision tokens after early decoding layers. SparseVLM~\citep{zhang2024sparsevlm} further introduces text-guided adaptive sparsification and recycles removed tokens into compact representations. MustDrop~\citep{liu2024mustdrop} progressively removes redundant tokens at multiple stages of the VLM pipeline. These methods reduce sequence length but depend on the reliability of their token-importance criteria. Another line of work focuses on similarity and duplication rather than individual token importance. ToMe~\citep{bolya2022tome} gradually merges similar tokens through lightweight matching. DART~\citep{wen2025dart} argues that duplication is a more reliable pruning signal and removes tokens that are highly similar to selected pivots. DivPrune~\citep{alvar2025divprune} likewise preserve representative or diverse visual information during token reduction. These approaches exploit visual redundancy directly, but altering the token set may still discard task-relevant details.

Sparse attention provides a complementary route that retains the tokens and KV cache while skipping less useful attention computation during prefilling. XAttention~\citep{xu2025xattention} estimates block importance using antidiagonal scores. MMInference~\citep{li2025mminference} exploits modality-specific and spatiotemporal grid patterns in long-context VLM attention. VecAttention~\citep{liu2026vecattention} further adopts fine-grained vector-wise sparsity for long-video understanding and generation. For Mixture-of-Experts VLMs, MoDES~\citep{huang2026modes} accelerates both prefilling and decoding through modality-aware dynamic expert skipping. These techniques reduce sequence-level or expert-level computation, whereas \method reduces model memory and linear-layer cost through low-bit quantization. The two directions are therefore orthogonal and can be combined, as demonstrated by the token-pruning experiments in Section~\ref{sec:experiments}.

\section{Preliminary}\label{sec:preliminary}

\begin{figure*}[!t]
    \centering
    \includegraphics[width=\linewidth]{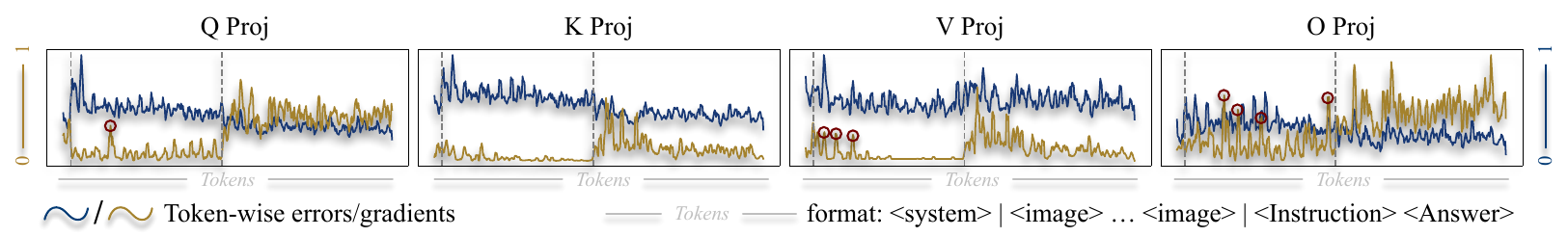}
    \caption{Visualization of normalized token-wise output-error and gradient magnitudes. The red circle indicates salient vision tokens. Output-error magnitudes remain relatively stable across tokens, whereas gradient magnitudes vary substantially across tokens and modalities.}
    \label{fig:err_grad}
\end{figure*}

\subsection{Transformer Model}
Modern Transformer-based VLMs~\citep{wang2024qwen2vl,bai2025qwen25vl,chen2024internvl2,li2024llava-ov} stack a series of decoding blocks, each formulated as
\begin{align}
    \rmY &= \rmX + \text{Attn}(\rmX) \label{eq:attnstream}, \\
    \rmZ &= \rmY + \text{FFN}(\rmY)   \label{eq:mlpstream},
\end{align}
where $\text{Attn}(\cdot)$ and $\text{FFN}(\cdot)$ denote the multi-head self-attention and feed-forward computations, and $\rmX, \rmY, \rmZ \in \sR^{\ci \times N}$ are the activations. Layer normalization is omitted for simplicity. Throughout this paper, we refer to $\text{Attn}(\cdot)$ and $\text{FFN}(\cdot)$ as \textit{blocks}, and to the linear projections within them (\emph{e.g.}, the query projection $\rmW_Q \in \sR^{\co \times \ci}$) as \textit{layers}.

\subsection{Quantization}
We follow the uniform affine quantization format widely used in LLM and VLM PTQ~\citep{frantar2022gptq,li2025gptqv2,li2025mbq}. For a linear layer with weight $\rmW \in \sR^{\co \times \ci}$ and activation $\rmX \in \sR^{\ci \times N}$, PTQ maps full-precision parameters to low-precision integers to reduce memory footprint and accelerate inference. Uniform affine quantization converts FP16/BF16 weights into INT-$B$ values through rounding and clipping:
\begin{align}
    &\text{Quant: } \rmW_\text{int}=\operatorname{Clamp}\left(\Bigl\lfloor\frac{\rmW}{s}\Bigr\rceil+z,\,0,\,2^B-1\right), \\
    &\text{Dequant: } \hat{\rmW}=s\times\left(\rmW_\text{int}-z\right),
\end{align}
where $\rmW_\text{int}$ denotes the quantized integer weight and $\hat{\rmW}$ is the pseudo-quantized weight in full precision. The step size $s$ and zero-point $z$ are computed from the dynamic range of $\rmW$ as
\begin{align}
    s&=\frac{\max\left(\rmW\right)-\min\left(\rmW\right)}{2^B-1}, \\
    z&=-\Bigl\lfloor\frac{\min\left(\rmW\right)}{s} \Bigr\rceil.
\end{align}
In our experiments, we apply symmetric ($z=0$) weight-only quantization to all linear layers in LLM backbones. When group-wise quantization is used, each weight group is quantized independently with its own scale $s$, and we set \texttt{group\_size}=128 unless otherwise specified. Existing PTQ methods calibrate quantization parameters by minimizing the layer-wise reconstruction error $\|\rmW\rmX - \hat{\rmW}\rmX\|_F^2$ over a small calibration set~\citep{frantar2022gptq,li2025gptqv2}. \method builds upon this calibration paradigm and incorporates token-level importance introduced in Section~\ref{sec:method}.

\subsection{Modality-Decomposed Reconstruction Objective}\label{sec:modality_decomposition}
For multimodal calibration, we partition the input activation matrix as $\rmX=\rmX_t\oplus\rmX_v$, where $\rmX_t$ and $\rmX_v$ contain the text and vision tokens, respectively. Let $\mathcal{T}$ and $\mathcal{V}$ denote their token-index sets, with sizes $N_t$ and $N_v$. The layer-wise output error is
\begin{equation}
    \Delta\rmZ=(\rmW-\hat{\rmW})\rmX.
\end{equation}
The conventional reconstruction objective can then be decomposed by modality:
\begin{align}
\gL_\text{rec}
&=\|\Delta\rmZ\|_F^2 \notag\\
&=\sum_{n\in\mathcal{T}}\|\Delta\rvz_n\|_2^2
 +\sum_{n\in\mathcal{V}}\|\Delta\rvz_n\|_2^2 \notag\\
&=N_t\bar{e}_t+N_v\bar{e}_v,
\label{eq:modality_decomposed_obj}
\end{align}
where $\Delta\rvz_n$ is the output error of token $n$, and $\bar{e}_t$ and $\bar{e}_v$ are the average per-token reconstruction errors of the two modalities:
\begin{equation}
\begin{aligned}
    \bar{e}_t
    &=\frac{1}{N_t}
    \sum_{n\in\mathcal{T}}\|\Delta\rvz_n\|_2^2,\\
    \bar{e}_v
    &=\frac{1}{N_v}
    \sum_{n\in\mathcal{V}}\|\Delta\rvz_n\|_2^2.
\end{aligned}
\end{equation}
Accordingly, the fraction of the reconstruction objective contributed by vision tokens is
\begin{equation}
    \rho_v=
    \frac{N_v\bar{e}_v}
    {N_t\bar{e}_t+N_v\bar{e}_v}.
\label{eq:vision_loss_ratio}
\end{equation}
In VLM calibration sequences, $N_v$ is often much larger than $N_t$. When the average token-wise errors are comparable, \emph{i.e.}, $\bar{e}_v\approx\bar{e}_t$, Eq.~\eqref{eq:vision_loss_ratio} approaches $N_v/(N_t+N_v)$. The objective is therefore dominated by vision tokens simply because they are more numerous. Normalizing the MSE by the total sequence length does not remove this relative imbalance. Moreover, many of these vision tokens are redundant and need not be equally important to the target loss. This decomposition exposes a fundamental mismatch between token-agnostic reconstruction and multimodal calibration. An effective objective should assign reconstruction effort according to token importance rather than token count.

\subsection{Motivation}\label{sec:motivation}
\begin{figure}[t]
\centering
\includegraphics[width=0.6\linewidth]{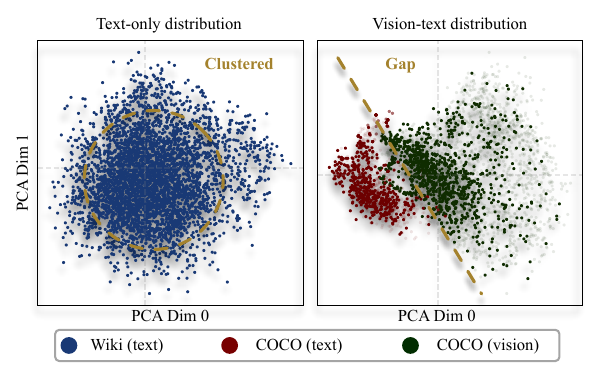} 
\caption{PCA visualization~\citep{shlens2014pca} of text-only~\citep{merity2016wikitext2} and vision-text~\citep{chen2024sharegpt4v} calibration data. Dark and light green points denote the important and unimportant vision tokens, respectively.}
\label{fig:hessian_variant}
\end{figure}

\subsubsection{Modality Discrepancy} VLMs jointly process text and vision inputs, but the two modalities are produced through different encoding pipelines and exhibit distinct activation statistics. Their token sequences differ not only in semantic content, but also in length, redundancy, and latent distribution. We characterize two PTQ-critical phenomena that jointly affect quantization performance:
\begin{itemize}[leftmargin=*, noitemsep, label=$\triangleright$]
    \item \textit{Observation 1 (visual over-representation)} Calibration sequences contain substantially more vision tokens than text tokens. Many vision tokens encode repeated local patterns or less informative background content. Their aggregate contribution can therefore dominate the reconstruction objective despite their limited task relevance (see Figure~\ref{fig:hessian_variant}).
    \item \textit{Observation 2 (modality gap)} Text and vision tokens occupy clearly separated regions of the latent feature space. This separation indicates that the two modalities follow different feature distributions and may respond differently to quantization perturbations (see Figure~\ref{fig:hessian_variant}).
\end{itemize}

These two phenomena reinforce each other. As shown by the modality decomposition in Eq.~\eqref{eq:modality_decomposed_obj}, the large number of vision tokens causes them to contribute most of the vanilla reconstruction loss. Meanwhile, the modality gap implies that this dominant visual contribution is not statistically equivalent to the contribution of text tokens. Token frequency therefore becomes a poor proxy for task relevance.

Existing LLM PTQ methods were primarily developed for text-only models. They minimize layer-wise mean squared error (MSE) while assigning equal weights to all tokens~\citep{lin2023awq,frantar2022gptq,shao2023omniquant}. When directly applied to VLMs, this token-agnostic paradigm overlooks both the numerical dominance and the heterogeneous importance of vision tokens. Consequently, calibration allocates excessive reconstruction effort to abundant but redundant visual features. Salient text and vision tokens receive insufficient emphasis, leading to notable performance degradation in low-bit settings (see Section~\ref{sec:experiments}).

\subsubsection{Hypothesis}
The modality-decomposed objective shows that token count can dominate calibration even when the average reconstruction errors of the two modalities are similar. This analysis suggests that equal weighting is not optimal for VLMs. We therefore hypothesize that reducing the contribution of redundant vision tokens can alleviate the visual-token bias. However, the weighting must remain selective. Moderate suppression should reduce redundancy, whereas excessive suppression may also weaken informative visual evidence. This hypothesis predicts that coarse down-weighting can be beneficial in favorable settings but will be sensitive to both the selected tokens and the weighting strength.

\subsubsection{Pilot Study}\label{sec:pilot}
We test this hypothesis on Qwen2-VL-7B-Instruct under INT2g128 quantization. We construct a naive importance-aware variant, denoted as SalQ-VLM$_\text{naive}$. It randomly marks a subset of vision tokens as low-importance (LI) and uniformly down-weights them during calibration. We vary both the LI ratio and the down-weighting factor to examine the sensitivity of this coarse strategy.

Table~\ref{tab:ablation_overall} reveals two important phenomena. First, random down-weighting can outperform an equal-weight baseline under suitable configurations. This result confirms that reducing the aggregate influence of vision tokens can mitigate calibration bias. Second, the improvement is inconsistent across weighting configurations and evaluation tasks. Increasing the LI ratio or suppression strength does not monotonically improve accuracy. This instability indicates that vision tokens are not uniformly redundant. A coarse modality-level rule may suppress useful visual information together with redundant content.

As a reference, Table~\ref{tab:ablation_overall} also reports the full gradient-driven \method introduced in the next section. It consistently achieves the best accuracy and exceeds the strongest naive configuration by more than two points on both benchmarks. The gap demonstrates that the key challenge is not merely reducing the total contribution of vision tokens. Instead, calibration requires a fine-grained signal that identifies which tokens should receive greater or lower emphasis. This controlled study therefore motivates the gradient-derived token importance factor developed in Section~\ref{sec:method}.

\begin{table}[!htbp]
\centering
\caption{Controlled Study of Coarse and Gradient-Driven Importance Strategies on Qwen2-VL-7B-Instruct Under INT2g128 Quantization.}
\label{tab:ablation_overall}
\resizebox{0.5\linewidth}{!}{\input{tables/ablation_overall}}
\end{table}

\section{Method}\label{sec:method}
Prior PTQ methods largely overlook visual over-representation and the modality gap in VLM activations. Motivated by these PTQ-critical yet underexplored characteristics, we propose \method, an importance-aware PTQ framework tailored to VLMs. To distinguish salient tokens from redundant ones, we introduce a diagonal gradient-driven importance factor $\rmG$, whose entries quantify the relative importance of individual tokens across modalities. This factor is theoretically grounded and experimentally validated. We compute $\rmG$ efficiently by caching raw gradients through a single lightweight block-wise backward pass at each decoding block, without updating model parameters. We then incorporate $\rmG$ into the layer-wise calibration objective to emphasize informative tokens while suppressing redundant vision tokens. In the remainder of this section, we first establish and derive the gradient-driven importance factor, then present the block-wise gradient caching scheme, and finally formulate the importance-aware calibration objective.

\subsection{Effective Gradient-Driven Importance Factor}
\subsubsection{Theoretically-grounded importance factor} Equation~\eqref{eq:modality_decomposed_obj} decomposes reconstruction error according to token count, but it does not measure how strongly each token affects the target loss. We therefore establish Theorem~\ref{thm:grad}, which connects the modality-decomposed reconstruction error to token-level loss sensitivity.
\begin{theorem}\label{thm:grad}
Let $\theta\in\sR^D$ denote the vectorized weights being quantized. Let $\rmZ(\theta)\in\sR^{\co\times N}$ be the corresponding layer output, and let $\rmP=\partial\gL/\partial\rmZ\in\sR^{\co\times N}$ be its gradient. A weight perturbation $\Delta\theta$ induces an output perturbation $\Delta\rmZ$. The resulting loss perturbation satisfies
\begin{align}
\Delta\gL
&=\left\langle\nabla_\theta\gL,\Delta\theta\right\rangle
+\gO\!\left(\|\Delta\theta\|_2^2\right) \notag\\
&=\left\langle\rmP,\Delta\rmZ\right\rangle_F
+\gO\!\left(\|\Delta\theta\|_2^2\right) \notag\\
&=\sum_{n\in\mathcal{T}}
\left\langle\rvp_n,\Delta\rvz_n\right\rangle
+\sum_{n\in\mathcal{V}}
\left\langle\rvp_n,\Delta\rvz_n\right\rangle
+\gO\!\left(\|\Delta\theta\|_2^2\right),
\label{eq:token_loss_perturbation}
\end{align}
where $\Delta\rvz_n$ and $\rvp_n$ are the $n$-th columns of $\Delta\rmZ$ and $\rmP$, respectively.
\end{theorem}
\begin{proof}
A first-order Taylor expansion around $\theta$ gives
\begin{equation}
\Delta\gL
=\sum_{i=1}^{D}
\Delta\theta_i\frac{\partial\gL}{\partial\theta_i}
+\gO\!\left(\|\Delta\theta\|_2^2\right).
\end{equation}
Applying the chain rule through the layer output yields
\begin{align}
\sum_{i=1}^{D}
\Delta\theta_i\frac{\partial\gL}{\partial\theta_i}
&=\sum_{i=1}^{D}\Delta\theta_i
\sum_{j=1}^{\co}\sum_{n=1}^{N}
\frac{\partial\gL}{\partial Z_{j,n}}
\frac{\partial Z_{j,n}}{\partial\theta_i} \notag\\
&=\sum_{j=1}^{\co}\sum_{n=1}^{N}
P_{j,n}
\left(
\sum_{i=1}^{D}
\Delta\theta_i
\frac{\partial Z_{j,n}}{\partial\theta_i}
\right) \notag\\
&=\sum_{j=1}^{\co}\sum_{n=1}^{N}
P_{j,n}\Delta Z_{j,n}
+\gO\!\left(\|\Delta\theta\|_2^2\right) \notag\\
&=\left\langle\rmP,\Delta\rmZ\right\rangle_F
+\gO\!\left(\|\Delta\theta\|_2^2\right).
\end{align}
Finally, partitioning the token index set into $\mathcal{T}$ and $\mathcal{V}$ gives Eq.~\eqref{eq:token_loss_perturbation}.
\end{proof}
The modality decomposition and Theorem~\ref{thm:grad} reveal two distinct measures of token contribution. Equation~\eqref{eq:modality_decomposed_obj} measures each token through its squared reconstruction error. In contrast, Eq.~\eqref{eq:token_loss_perturbation} measures its first-order effect on the target loss. For clarity, we define the magnitude of this effect and its visual-modality share as
\begin{equation}
\begin{aligned}
c_n&=\left|\left\langle\rvp_n,\Delta\rvz_n\right\rangle\right|,\\
\eta_v&=
\frac{\sum_{n\in\mathcal{V}}c_n}
{\sum_{n\in\mathcal{T}}c_n+\sum_{n\in\mathcal{V}}c_n}.
\end{aligned}
\end{equation}
The reconstruction share $\rho_v$ in Eq.~\eqref{eq:vision_loss_ratio} and the task-sensitivity share $\eta_v$ need not agree. The former grows with the number and error magnitude of vision tokens. The latter also depends on their gradients. Therefore, numerous vision tokens can dominate vanilla MSE while contributing little to the target loss. Tokens with similar reconstruction errors can likewise have very different task impacts.

Figure~\ref{fig:err_grad} empirically supports this distinction. Token-wise output-error magnitudes remain relatively similar, whereas gradient magnitudes vary substantially. Most vision tokens have small gradients, while a small subset exhibits large gradients. The token-count-dominated MSE therefore overemphasizes many low-sensitivity vision tokens and underemphasizes a few salient tokens. Since the output errors provide limited discrimination in this setting, we use gradient statistics as a practical proxy for token-level loss sensitivity. This choice directly connects the modality-decomposed objective to the gradient-driven importance factor defined below.

\subsubsection{Gradient Processing} Theorem~\ref{thm:grad} measures the first-order contribution of token $n$ by $c_n=|\langle\rvp_n,\Delta\rvz_n\rangle|$. By the Cauchy-Schwarz inequality,
\begin{equation}
    c_n
    \leq
    \|\rvp_n\|_2\,\|\Delta\rvz_n\|_2.
\label{eq:token_sensitivity_bound}
\end{equation}
The output error $\Delta\rvz_n$ depends on the quantized weight candidate and changes during layer-wise optimization. In contrast, the gradient $\rvp_n$ can be cached once and reused throughout calibration. Figure~\ref{fig:err_grad} also shows that output-error magnitudes are relatively similar across tokens, while gradient magnitudes provide substantially stronger discrimination. We therefore use gradient magnitude as a practical proxy for token-level loss sensitivity.

For a given layer, the raw output gradient is $\rmP\in\sR^{\co\times N}$, and its $n$-th column is $\rvp_n$. We aggregate the absolute gradient over output channels:
\begin{equation}\label{eq:grad2score}
\begin{aligned}
g_n
&=\frac{1}{\co}\|\rvp_n\|_1
=\frac{1}{\co}\sum_{i=1}^{\co}|P_{i,n}|,\\
\rmG
&=\operatorname{Diag}\left(g_1,g_2,\ldots,g_N\right).
\end{aligned}
\end{equation}
The channel average prevents the score magnitude from scaling with the output dimension $\co$. Each diagonal entry of $\rmG$ then weights one token during calibration. Tokens with larger gradients receive greater emphasis, whereas low-sensitivity vision tokens are naturally down-weighted. Multiplying all $g_n$ by the same positive constant only rescales the objective and does not change its minimizer.

This construction operationalizes the distinction between the reconstruction share $\rho_v$ and task-sensitivity share $\eta_v$ introduced earlier. Vanilla MSE emphasizes tokens according to their error and frequency. In contrast, $\rmG$ redistributes calibration effort according to loss sensitivity. We also evaluate attention-score-based alternatives~\citep{chen2024fastv,dhouib2025pact}. Such scores estimate relevance for token pruning rather than sensitivity to quantization perturbations. They consequently underperform the proposed gradient factor, as shown in Section~\ref{sec:ablation}.

\begin{figure}[t]
    \centering
    \includegraphics[width=0.6\linewidth]{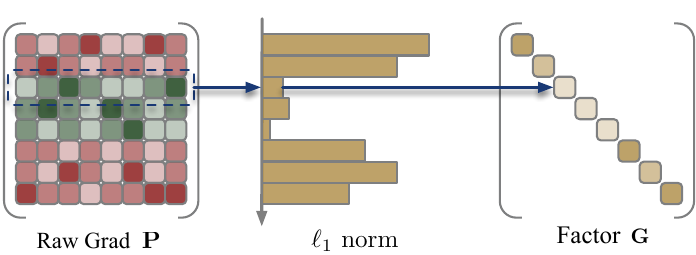} 
    \caption{Derivation of the gradient-driven importance factor $\rmG$. Absolute values in each token-gradient column of $\rmP$ are averaged over output channels to obtain $g_n$, which forms the corresponding diagonal entry of $\rmG$.}
    \label{fig:derive_visual_factor}
\end{figure}

\subsection{Efficient Block-wise Gradient Caching}
To instantiate Theorem~\ref{thm:grad}, we require a target loss from which the output gradients can be obtained. We consider three levels of supervision: a layer-wise distillation loss, a block-wise distillation loss, and a network-wise supervised fine-tuning loss. The layer-wise loss is inexpensive but ignores interactions among projections within a Transformer block~\citep{li2021brecq}. The network-wise loss captures global dependencies but requires backpropagation through the full model. It is therefore expensive and may overfit the limited calibration set~\citep{gong2025pushing}. We adopt block-wise supervision as a balance between these extremes.

\begin{figure*}[!t]
  \centering
  \includegraphics[width=\linewidth]{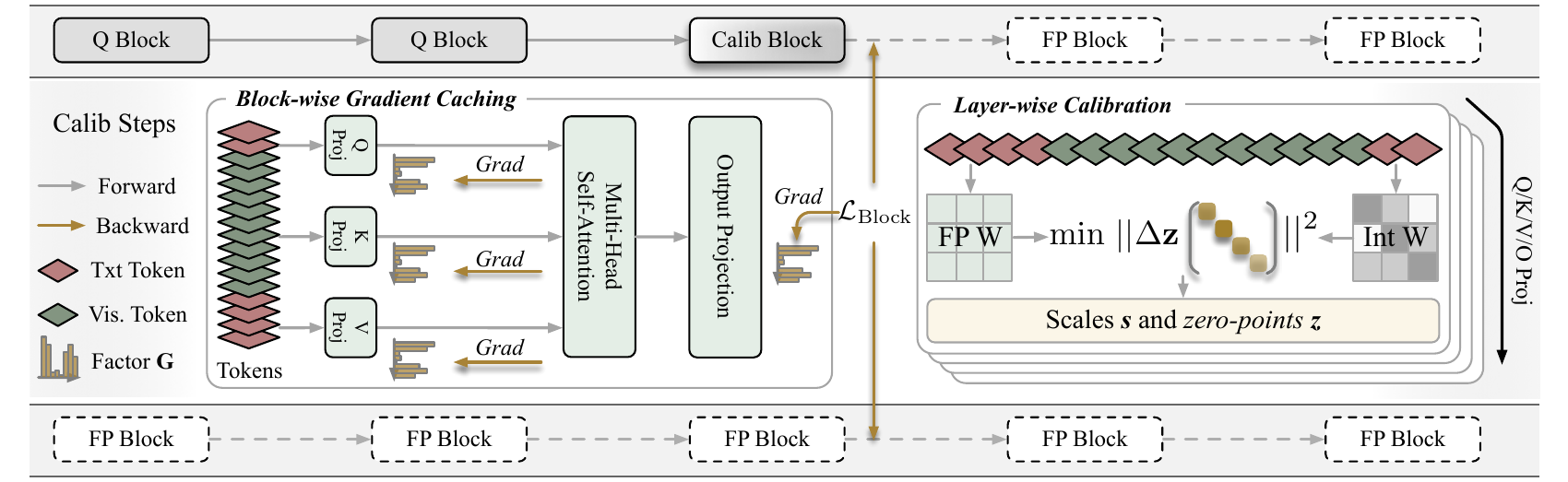} 
  \caption{Overview of \method. \textbf{Left:} Block-wise gradient caching. The discrepancy between full-precision and semi-quantized block outputs defines $\gL_\text{Block}$. One local backward pass caches projection-output gradients and derives layer-specific importance factors. \textbf{Right:} Importance-aware layer-wise calibration. Layers are quantized sequentially using the $\rmG$-weighted objective. The two stages alternate across decoding blocks until the full model is quantized.}
  \label{fig:vlmq_overview}
\end{figure*}

Consider decoding block $b$. Let $\rmY_b^\text{FP}$ be its full-precision output and $\rmY_b^\text{SQ}$ be the output on the semi-quantized path. The latter receives activations produced by all previously quantized layers. We define the local residual-stream objective as
\begin{equation}\label{eq:loss_func}
    \gL_\text{Block}^{(b)}
    =
    \left\|
    \rmY_b^\text{FP}-\rmY_b^\text{SQ}
    \right\|_F^2.
\end{equation}
For each linear layer $\ell$ in block $b$, let $\rmZ^{(\ell)}$ denote its output activation. The raw gradient used by Eq.~\eqref{eq:grad2score} is
\begin{equation}
    \rmP^{(\ell)}
    =
    \frac{\partial\gL_\text{Block}^{(b)}}
    {\partial\rmZ^{(\ell)}}.
\label{eq:block_gradient}
\end{equation}
As illustrated in Figure~\ref{fig:vlmq_overview} (left), a single backward pass through the current block produces the gradients for all enhanced projections. We cache only the resulting $\rmP^{(\ell)}$ tensors and release the computation graph before quantizing the block. This localized procedure captures intra-block dependencies while bounding memory and computation to one block at a time. It also requires no parameter update. The comparison in Section~\ref{sec:ablation} confirms that this granularity provides the best accuracy-efficiency trade-off.

\subsection{Importance-Aware Layer-Wise Calibration}
As shown in Figure~\ref{fig:vlmq_overview} (right), we incorporate the layer-specific importance matrix $\rmG$ into the reconstruction objective. Let $\mathcal{Q}_B$ denote the feasible set of weights under the target $B$-bit quantization format. The calibrated quantized weight is
\begin{align}
\rmW_\text{Int}^{\star}
={}&\arg\min_{\rmW_\text{Int}\in\mathcal{Q}_B}
\left\|
\left(
\rmW_\text{FP}\rmX-\rmW_\text{Int}\rmX
\right)\rmG
\right\|_F^2.
\label{eq:vlmq_obj}
\end{align}
Because $\rmG$ is diagonal, this objective can be written as a token-weighted and modality-decomposed reconstruction loss:
\begin{align}
\gL_\text{IA}
={}&
\sum_{n\in\mathcal{T}}
g_n^2
\left\|
(\rmW_\text{FP}-\rmW_\text{Int})\rmX_{:,n}
\right\|_2^2 \notag\\
&+
\sum_{n\in\mathcal{V}}
g_n^2
\left\|
(\rmW_\text{FP}-\rmW_\text{Int})\rmX_{:,n}
\right\|_2^2.
\label{eq:importance_modality_obj}
\end{align}
Equation~\eqref{eq:importance_modality_obj} directly refines Eq.~\eqref{eq:modality_decomposed_obj}. The vanilla objective weights every token equally and is therefore dominated by $N_v$. Our objective instead scales token $n$ by $g_n^2$. Salient tokens contribute more strongly, while abundant low-gradient vision tokens contribute less. The square arises naturally because $\rmG$ is applied inside the Frobenius norm.

The proposed formulation changes how reconstruction errors are aggregated across tokens. It does not prescribe a particular quantization solver. Existing layer-wise PTQ pipelines can retain their original quantizer, search strategy, and weight-update procedure while replacing the uniform reconstruction loss with Eq.~\eqref{eq:vlmq_obj}. A non-uniform $\rmG$ instead reallocates reconstruction effort toward tokens with greater loss sensitivity. The proposed contribution is therefore an importance-aware optimization objective that is independent of a specific precursor algorithm.

To quantize the full model, \method processes decoding blocks sequentially. At each block, it first computes $\gL_\text{Block}$ and caches $\rmP^{(\ell)}$. It then derives $\rmG^{(\ell)}$ and calibrates the corresponding layers in order. The resulting quantized outputs become the inputs to the next block. This progressive alternation allows later importance factors to reflect errors propagated from preceding quantized layers. Since $\rmG$ is used only during calibration and is not retained by the deployed model, \method introduces no additional inference-time operation.

\section{Importance-Aware Formulation}\label{appendix:vlmq_framework}
We instantiate \method on GPTAQ~\citep{li2025gptqv2}, which serves as our default precursor. GPTAQ minimizes asymmetric reconstruction error between the full-precision and quantized activation paths. However, its objective assigns equal importance to all output tokens in $\rmZ=\rmW\rmX$. This uniform treatment can bias calibration toward the numerous redundant vision tokens.

Following GPTAQ, let $\rmX$ and $\rmX_\text{FP}$ denote the quantized-path and full-precision-path input activations, respectively. For a weight row $\rvw$, the output residual is
\begin{equation}
    \rvr=\rvw\rmX_\text{FP}-\rvw\rmX.
\end{equation}
We introduce a diagonal token-level importance matrix $\rmG\in\sR^{N\times N}$. Its $i$-th diagonal entry specifies the importance of the $i$-th output token. When quantizing the $q$-th weight, the importance-aware constrained objective is
\begin{equation}\label{eq:refined_gptaq_obj}
\begin{aligned}
    \Delta\rvw^\star
    ={}&\arg\min_{\Delta\rvw}
    \left\|\left(\Delta\rvw\rmX-\rvr\right)\rmG\right\|_2^2,\\
    &\text{s.t.}\quad
    \Delta\rvw\rve_q^\top+\rvw_q-\hat{\rvw}_q=0,
\end{aligned}
\end{equation}
where $\rve_q$ is the one-hot vector whose $q$-th entry is one.

For compactness, we define the importance-aware activation, residual, and Hessian as
\begin{equation}
\begin{aligned}
\widetilde{\rmX}&=\rmX\rmG,
\widetilde{\rvr}=\rvr\rmG,\\
\widetilde{\rmH}
&=\widetilde{\rmX}\widetilde{\rmX}^{\top}
=\rmX\rmG^2\rmX^\top.
\end{aligned}
\end{equation}
The Lagrangian of \eqref{eq:refined_gptaq_obj} is then
\begin{equation}\label{eq:lagrangian}
\begin{aligned}
\mathcal{J}={}&
\left\|\Delta\rvw\widetilde{\rmX}-\widetilde{\rvr}\right\|_2^2\\
&+\mu\left(
\Delta\rvw\rve_q^\top+\rvw_q-\hat{\rvw}_q
\right),
\end{aligned}
\end{equation}
where $\mu$ is the Lagrange multiplier. Setting the derivatives with respect to $\Delta\rvw$ and $\mu$ to zero gives
\begin{equation}
\begin{cases}
2\Delta\rvw\widetilde{\rmH}
-2\widetilde{\rvr}\widetilde{\rmX}^{\top}
+\mu\rve_q=0,\\
\Delta\rvw\rve_q^\top+\rvw_q-\hat{\rvw}_q=0.
\end{cases}
\end{equation}
The first condition can be rearranged as
\begin{equation}
\Delta\rvw\widetilde{\rmH}
=-\frac{\mu}{2}\rve_q
+\widetilde{\rvr}\widetilde{\rmX}^{\top}.
\end{equation}
Enforcing the quantization constraint yields the importance-aware GPTAQ update
\begin{equation}\label{eq:vlmq_weight_updating}
\begin{aligned}
\Delta\rvw^\star
={}&
\frac{\hat{\rvw}_q-\rvw_q}
{\left[\widetilde{\rmH}^{-1}\right]_{qq}}
\left[\widetilde{\rmH}^{-1}\right]_{q,:}\\
&+\widetilde{\rvr}\widetilde{\rmX}^{\top}
\widetilde{\rmH}_{-q}^{-1},
\end{aligned}
\end{equation}
where $\widetilde{\rmH}_{-q}^{-1}$ denotes the inverse Hessian after removing the $q$-th row and column. Equation~\eqref{eq:vlmq_weight_updating} preserves the closed-form structure of GPTAQ after replacing the original activation, residual, and Hessian with their importance-aware counterparts. Unlike BoA~\citep{kim2024boa}, which assigns importance at the channel level, \method operates at token-level activation granularity. The weighting mechanism is also orthogonal to the parallelization strategies of GPTQ~\citep{frantar2022gptq} and GPTAQ~\citep{li2025gptqv2}. Therefore, \method can directly inherit their implementation optimizations, including Cholesky-based reformulation.

\section{Experiments}\label{sec:experiments}
\subsection{Implementation Details}
\subsubsection{Models} We evaluate \method on three representative VLM backbones: LLaVA-OneVision-7B~\citep{li2024llava-ov}, Qwen2-VL-7B-Instruct~\citep{wang2024qwen2vl}, and Qwen2.5-VL-7B-Instruct~\citep{bai2025qwen25vl}. We further assess its scalability on Qwen2-VL-2B-Instruct and Qwen2.5-VL-32B-Instruct, covering model sizes from 2B to 32B parameters.

\subsubsection{Configuration} We construct the calibration set by randomly sampling 512 image-text pairs from the improved COCO Caption dataset provided by ShareGPT4V~\citep{chen2024sharegpt4v}. By default, \method is built upon GPTAQ~\citep{li2025gptqv2}, with the corresponding importance-aware formulation detailed in Section~\ref{appendix:vlmq_framework}. Unless otherwise specified, we set the dampening ratio to $\lambda=0.01$ and enable \texttt{act\_order}. We use Flash Attention~\citep{dao2022flashattention} during evaluation and retain the default settings for all remaining configurations.

\subsubsection{Benchmarks} We evaluate quantized VLMs using the LMMs-Eval framework~\citep{zhang2024lmms-eval} on ChartQA~\citep{masry2022chartqa}, DocVQA~\citep{mathew2021docvqa}, MME-RealWorld~\citep{zhang2024mme-realword} in both English and Chinese, OCRBench~\citep{liu2024ocrbench}, ScienceQA~\citep{lu2022scienceqa}, SeedBench 2 Plus~\citep{li2024seed2plus}, and TextVQA~\citep{singh2019textvqa}. Together, these eight evaluations measure text recognition, visual perception, and visual reasoning. To examine generalization beyond standard vision-language benchmarks, we additionally evaluate on the text-only HellaSwag~\citep{zellers2019hellaswag}, the text-heavy TextCaps~\citep{sidorov2020textcaps}, the multi-image reasoning benchmark MMMU~\citep{yue2023mmmu}, and the long-video benchmarks VideoMME~\citep{fu2025videomme} and LongVideoBench~\citep{wu2024longvideobench}.

\subsubsection{Baselines} We compare \method with four representative training-free PTQ methods. AWQ~\citep{lin2023awq} applies equivalent transformations based on salient input channels. MBQ~\citep{li2025mbq} adapts AWQ to VLMs by accounting for the different quantization sensitivities of text and vision tokens. GPTQ~\citep{frantar2022gptq} uses second-order information to minimize layer-wise reconstruction error. GPTAQ~\citep{li2025gptqv2} further employs asymmetric calibration to account for accumulated quantization errors.

\subsection{Main Results}

\subsubsection{INT3 quantization}
\begin{table*}[!t]
\centering
\caption{INT3 quantization results on eight vision-language benchmarks.}
\label{tab:int3_exp}
\resizebox{0.95\linewidth}{!}{\input{tables/int3_exp}}
\end{table*}
Table~\ref{tab:int3_exp} reports the results under ungrouped INT3 quantization. This setting exposes clear differences among PTQ paradigms. The smoothing-based AWQ and MBQ largely collapse on the Qwen2-VL models, whereas the Hessian-based methods remain substantially more stable. Impressively, \method achieves the highest average accuracy in all five model settings. Its gains are consistent across Qwen2-VL, Qwen2.5-VL, and LLaVA-OneVision at the 7B scale, indicating that token-level importance-aware calibration is not specific to one VLM architecture. The improvement is especially visible on quantization-sensitive multimodal reasoning tasks. For example, \method improves MME-RealWorld (Chinese) by 3.29 points over GPTAQ on Qwen2-VL-7B-Instruct. Model scale also affects the improvement margin. The compact 2B model obtains the largest gain over the strongest baseline but still remains far from full precision, showing greater vulnerability to low-bit errors. In contrast, the 32B model is inherently more robust and leaves less headroom, yet \method still retains the best average performance. Overall, the results show that \method provides consistent aggregate gains, with its advantage becoming stronger when quantization-induced calibration bias is more severe.

\subsubsection{INT3g128 quantization}
\begin{table*}[!t]
\centering
\caption{INT3g128 quantization results on eight vision-language benchmarks.}
\label{tab:int3g128_exp}
\resizebox{0.95\linewidth}{!}{\input{tables/int3g128_exp}}
\end{table*}
Table~\ref{tab:int3g128_exp} evaluates group-wise INT3 quantization with \texttt{group\_size=128}. Group-wise quantization reduces the difficulty of weight reconstruction, so the strongest baselines already remain close to full precision. Consequently, the improvement margins are narrower than those under ungrouped INT3. Nevertheless, \method achieves the highest average accuracy on all three backbones. Its benefit is most visible on quantization-sensitive tasks. For example, it improves MME-RealWorld (Chinese) from 52.12\% with GPTAQ to 55.47\% on Qwen2-VL-7B-Instruct. This trend shows that token-level importance remains useful even when the overall quantization error is moderate.

\subsubsection{INT2g128 quantization}
\begin{table*}[!t]
\centering
\caption{INT2g128 quantization results on eight vision-language benchmarks.}
\label{tab:int2g128_exp}
\resizebox{0.95\linewidth}{!}{\input{tables/int2g128_exp}}
\end{table*}
Table~\ref{tab:int2g128_exp} further evaluates the more challenging INT2g128 setting. At this precision, quantization errors are strongly amplified and the differences among calibration strategies become more pronounced. AWQ~\citep{lin2023awq} and MBQ~\citep{li2025mbq} collapse on Qwen2-VL-7B-Instruct, while GPTAQ becomes unstable on Qwen2.5-VL-7B-Instruct. \method achieves the highest average accuracy on all three backbones despite these failures. The strongest recovery appears on MME-RealWorld (Chinese), where it improves Qwen2.5-VL-7B-Instruct by 16.45 points over GPTQ. These results show that importance-aware calibration becomes increasingly valuable as the quantization error grows.

\subsubsection{Generalization to more tasks}
Table~\ref{tab:video_text} examines whether the calibration benefit transfers beyond standard VQA. We evaluate the same INT2g128 Qwen2-VL-7B-Instruct model without task-specific recalibration. The tasks span text-only reasoning, visually grounded generation, multi-image reasoning, and long-video understanding. \method outperforms both GPTQ and GPTAQ throughout this diverse evaluation. The advantage also persists for long multimodal sequences. On LongVideoBench, for example, \method exceeds the strongest baseline by 2.08 points. This consistency indicates that the \method captures general token-importance patterns rather than overfitting to the calibration task.

\begin{table}[!htbp]
\centering
\caption{Generalization beyond standard VQA benchmarks. All quantized results are obtained with Qwen2-VL-7B-Instruct under INT2g128 quantization without task-specific recalibration.}
\label{tab:video_text}
\resizebox{0.5\linewidth}{!}{\input{tables/video_text_results}}
\end{table}

\subsubsection{Synergy with token pruning} \method is orthogonal to token compression and can be combined with existing pruning techniques. We integrate INT2g128 \method with FastV~\citep{chen2024fastv} and ToMe~\citep{bolya2022tome} at several pruning ratios. Table~\ref{tab:pruning_comparison} shows that \method consistently outperforms GPTQ after pruning and remains robust under aggressive token removal. More importantly, three of the four pruned variants also surpass the unpruned \method baseline. For example, ToMe at 80\% pruning improves MMMU accuracy from 35.11\% to 35.78\%. Importance-aware quantization reduces the influence of redundant tokens during calibration, whereas pruning removes redundant computation during inference. Their complementary mechanisms therefore produce a notable $1{+}1{>}2$ effect.

\begin{table}[!htbp]
\centering
\caption{MMMU Accuracy With FastV and ToMe at Various Pruning Ratios Under INT2g128 Quantization.}
\label{tab:pruning_comparison}
\resizebox{0.65\linewidth}{!}{\input{tables/pruning_comparison}}
\end{table}

\subsection{Ablation Study}\label{sec:ablation}
\subsubsection{Ablation on importance factor type}
Attention scores from FastV~\citep{chen2024fastv} and PACT~\citep{dhouib2025pact} provide alternative measures of token significance. Table~\ref{tab:ablation_importance_factor_type_compact} compares them with constant weighting and the proposed gradient-driven factor under the same INT3 setting. Only the gradient-derived factor improves upon constant weighting. Both attention-based factors instead reduce quantized accuracy. The largest gap is observed against PACT, where \method gains 2.63 points. Thus, token relevance for pruning does not directly represent sensitivity to quantization error. Gradients provide a better-aligned importance indicator for importance-aware PTQ.

\begin{table}[!htbp]
\centering
\caption{Ablation Study of Importance-Factor Types on Qwen2-VL-7B-Instruct Under INT3 Quantization.}
\label{tab:ablation_importance_factor_type_compact}
\resizebox{0.6\linewidth}{!}{\input{tables/ablation_importace_factor_type_compact}}
\end{table}

\subsubsection{Ablation on backpropagation granularity} 
We next compare layer-wise, block-wise, and network-wise objectives for acquiring raw gradients. Table~\ref{tab:ablation_back_granularity_compact} shows that block-wise gradient caching provides the best overall trade-off. It achieves the highest accuracy and the shortest quantization time, while matching the low memory cost of the network-wise alternative. In the representative INT3 setting, it reaches 69.70\% average accuracy in 0.21 GPU hours with approximately 30 GB of memory. The layer-wise objective ignores cross-layer dependencies and yields less informative gradients. The network-wise objective captures broader dependencies but is substantially slower and may overfit the limited calibration set. Block-wise backpropagation therefore offers the most effective and practical granularity for \method.

\begin{table}[!htbp]
\centering
\caption{Ablation Study of Gradient-Caching Granularity Under INT3 Quantization.}
\label{tab:ablation_back_granularity_compact}
\resizebox{0.6\linewidth}{!}{\input{tables/ablation_back_granularity_compact}}
\end{table}

\subsubsection{Ablation on layer selection} 
We also examine the local loss and the set of enhanced attention projections. Table~\ref{tab:ablation_layer_selection_compact} shows that enhancing only the query, key, and value projections causes severe degradation under both tested losses. Changing the loss alone therefore cannot compensate for incomplete layer coverage. In contrast, the configuration that applies the residual-stream loss $||\Delta\rmY||_2^2$ to all query, key, value, and output projections restores the average accuracy to 69.70\%. The dominant observation is that \texttt{o\_proj} must be included to preserve error propagation through the complete attention block. We therefore use the residual-stream objective and enhance all four projections in \method.

\begin{table}[!htbp]
\centering
\caption{Ablation Study of Local Loss Terms and Enhanced Layers Under INT3 Quantization.}
\label{tab:ablation_layer_selection_compact}
\resizebox{0.5
\linewidth}{!}{\input{tables/ablation_layer_selection_compact}}
\end{table}

\subsubsection{Ablation on precursor algorithms} 
A key advantage of \method is its plug-and-play compatibility with existing PTQ algorithms. The importance-aware objective can be added without redesigning the underlying quantization pipeline. Table~\ref{tab:ablation_precursor} shows consistent improvements when \method is applied to GPTQ and GPTAQ across both model families. Its benefit is especially large when GPTAQ becomes unstable under INT2. On Qwen2.5-VL-7B-Instruct, for example, \method recovers 10.41 points. However, \method does not rescue AWQ after the smoothing-based precursor has already collapsed. Plug-and-play compatibility therefore broadens the applicability of \method, but still requires a viable precursor.

\begin{table}[!htbp]
\centering
\caption{Average Accuracy of Different Precursors With and Without \method Under INT2g128 Quantization.}
\label{tab:ablation_precursor}
\resizebox{0.5\linewidth}{!}{\input{tables/ablation_precursor}}
\end{table}

\subsection{Efficiency}
\subsubsection{Quantization efficiency}
\begin{figure}[t]
\centering
\includegraphics[width=0.6\linewidth]{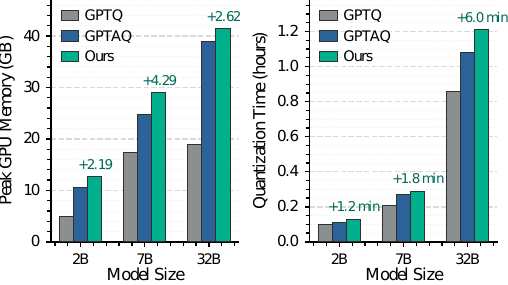}
\caption{Quantization Cost Across Model Scales on a Single H100 80GB GPU. Labels above the \method bars report the additional memory or time relative to GPTAQ.}
\label{fig:quant_efficiency}
\end{figure}
Figure~\ref{fig:quant_efficiency} compares quantization time and peak GPU memory from 2B to 32B parameters. The additional cost of \method remains limited across all model scales. Relative to GPTAQ, the maximum overhead is 4.29 GB of peak memory and 6 minutes of quantization time. This cost mainly comes from temporarily storing activation gradients for one local backward pass per block. The overhead does not prevent large-model calibration. In particular, the 32B model remains quantizable on a single H100 80GB GPU. \method therefore scales to large VLMs while retaining a practical calibration budget.

\subsubsection{Inference speedup}
\method modifies only the calibration objective and preserves the standard GPTQ-compatible W3 representation~\citep{frantar2022gptq}. Its quantized weights can therefore be executed directly by existing hardware-optimized kernels without additional runtime modules. Table~\ref{tab:speedup} reports the latency of representative Qwen2-VL-7B-Instruct linear layers on an RTX 5090 GPU. Across both attention and MLP projections, the W3 kernels consistently deliver over $9\times$ speedups over FP16. These results confirm that \method inherits the inference acceleration of the underlying GPTQ deployment pipeline without introducing extra inference overhead.

\begin{table}[!htbp]
\centering
\caption{Kernel Latency ($\mu$s) of Representative Qwen2-VL-7B-Instruct Linear Layers in FP16 and W3. Weight dimensions follow $\texttt{in\_dim}\times\texttt{out\_dim}$.}
\label{tab:speedup}
\resizebox{0.6\linewidth}{!}{\input{tables/speedup}}
\end{table}

\subsubsection{Compatibility with hardware-optimized kernels} The unchanged weight representation also makes \method compatible with the broader GPTQ deployment ecosystem. Specialized backends such as Marlin~\citep{frantar2024marlin} and ExLLaMA~\citep{exllamav2} can be used without kernel modification or method-specific hardware support. Thus, the importance-aware calibration procedure remains independent of the inference implementation. \method can immediately benefit from future improvements to GPTQ-compatible kernels.

\section{Conclusion and Limitations}\label{sec:conclusion}
We presented \method, an importance-aware PTQ framework tailored to VLMs. We characterized visual over-representation and the modality gap as two overlooked challenges in VLM quantization. These properties create a mismatch between redundant vision tokens and conventional token-agnostic calibration. To address this mismatch, \method derives token-level importance from output gradients. The factor is obtained through a single lightweight block-wise backward pass and incorporated into the layer-wise reconstruction objective. This design prioritizes salient tokens while suppressing redundant ones. It is plug-and-play with existing layer-wise PTQ pipelines and preserves compatibility with existing low-bit inference kernels.

Extensive evaluations cover VLMs from 2B to 32B parameters, eight primary benchmarks, and five additional generalization tasks. \method consistently achieves the highest average accuracy across the evaluated quantization settings. Its advantage is particularly pronounced under ultra-low-bit quantization. On MME-RealWorld, it improves accuracy by 16.45 percentage points under INT2g128. The method also transfers to text-only, multi-image, and long-video tasks. Moreover, it remains complementary to token pruning and introduces only limited calibration overhead.

Our current study focuses on symmetric weight-only quantization of linear layers in the LLM backbone. Quantizing the vision encoder, multimodal projector, and activations remains unexplored. Future work will extend \method to full-model weight-activation quantization, broader video-oriented VLMs, and longer temporal contexts.

\bibliographystyle{plainnat}
\bibliography{paper}

\end{document}

%% file: math_commands.tex
\usepackage{amsmath,amsfonts,bm}

\def\eqref#1{equation~\ref{#1}}

\def\1{\bm{1}}

\def\rve{{\mathbf{e}}}

\def\rvp{{\mathbf{p}}}

\def\rvr{{\mathbf{r}}}

\def\rvw{{\mathbf{w}}}

\def\rvz{{\mathbf{z}}}

\def\rmG{{\mathbf{G}}}
\def\rmH{{\mathbf{H}}}

\def\rmP{{\mathbf{P}}}

\def\rmW{{\mathbf{W}}}
\def\rmX{{\mathbf{X}}}
\def\rmY{{\mathbf{Y}}}
\def\rmZ{{\mathbf{Z}}}

\DeclareMathAlphabet{\mathsfit}{\encodingdefault}{\sfdefault}{m}{sl}
\SetMathAlphabet{\mathsfit}{bold}{\encodingdefault}{\sfdefault}{bx}{n}

\def\gL{{\mathcal{L}}}

\def\gO{{\mathcal{O}}}

\def\sR{{\mathbb{R}}}

%% file: tables/ablation_overall.tex
\begin{tabular}{
  >{\raggedright\arraybackslash}m{2.5cm}
  *{2}{>{\centering\arraybackslash}m{1.05cm}}
  *{2}{>{\centering\arraybackslash}m{1.15cm}}
}
\toprule
\multirow{2}{*}{\textbf{Method}}
& \multicolumn{2}{c}{\textbf{Importance Strategy}}
& \multicolumn{2}{c}{\textbf{Accuracy} ($\uparrow$)} \\
\cmidrule(lr){2-3}\cmidrule(lr){4-5}
& \textbf{LI Ratio} & $\gamma$
& \textbf{TextVQA} & \textbf{DocVQA} \\
\midrule
GPTQ  & - & - & 71.93 & \underline{74.90} \\
\midrule
\multirow{9}{*}{SalQ-VLM$_{\text{naive}}$}
& \multirow{3}{*}{0.25} & 0.01 & 70.66 & 72.12 \\
&                         & 0.05 & 72.12 & 72.63 \\
&                         & 0.10 & 70.89 & 72.10 \\
\cmidrule(lr){2-5}
& \multirow{3}{*}{0.50} & 0.01 & \underline{72.38} & 72.76 \\
&                         & 0.05 & 69.83 & 71.93 \\
&                         & 0.10 & 71.79 & 73.67 \\
\cmidrule(lr){2-5}
& \multirow{3}{*}{0.75} & 0.01 & 69.74 & 69.44 \\
&                         & 0.05 & 68.17 & 71.05 \\
&                         & 0.10 & 71.21 & 70.33 \\
\midrule
\rowcolor{bgblue!10}\method (Ours)
& \multicolumn{2}{c}{Gradient-driven}
& \textbf{74.80} & \textbf{75.76} \\
\bottomrule
\end{tabular}

%% file: tables/int3_exp.tex
\begin{tabular}{>{
    \raggedright\arraybackslash}m{2.0cm}
    *{8}{>{\centering\arraybackslash}m{1.2cm}}
    >{\raggedright\arraybackslash}m{1.5cm}
}
\toprule
\textbf{Method} & \textbf{Chart} & \textbf{Doc}$^\text{val}$ & \textbf{MME}$_\text{en}$ & \textbf{MME}$_\text{cn}$ & \textbf{OCR} & \textbf{SciQA} & \textbf{Seed}$^\text{2+}$ & \textbf{Text}$^\text{val}$ & \textbf{Avg} $(\uparrow)$\\
\midrule

\multicolumn{10}{c}{\textit{Qwen2-VL-2B-Instruct-INT3}} \\
\midrule
Full Prec & 72.61 & 89.35 & 40.27 & 43.59 & 76.60 & 74.23 & 61.79 & 79.38 & 67.23 \\
AWQ & 5.36 & 10.02 & 9.65 & 8.48 & 5.40 & 1.37 & 18.53 & 4.37 & 7.90 \\
MBQ & 3.56 & 10.19 & 10.22 & 7.79 & 4.40 & 1.44 & 8.12 & 3.90 & 6.20 \\
GPTQ  & 58.84 & 74.55 & 33.25 & 30.83 & 61.50 & 65.13 & 50.37 & 73.49 & 56.00 \\
GPTAQ & 58.76 & 74.03 & 38.71 & 28.71 & 61.10 & 64.25 & 52.17 & 73.18 & 56.36 \\
\rowcolor{bgblue!10}\method & 59.44 & 76.59 & 37.94 & 32.58 & 61.90 & 63.05 & 52.92 & 74.54 & \textbf{57.37}$_\textbf{(+1.01\%)}$ \\
\midrule

\multicolumn{10}{c}{\textit{Qwen2-VL-7B-Instruct-INT3}} \\
\midrule
Full Prec & 81.44 & 93.89 & 57.30 & 56.09 & 80.90 & 85.55 & 69.26 & 82.02 & 75.81 \\
AWQ & 18.20 & 27.67 & 30.20 & 20.03 & 8.90 & 49.89 & 44.88 & 28.23 & 28.50 \\
MBQ & 16.52 & 24.75 & 27.23 & 21.35 & 8.10 & 35.25 & 40.10 & 27.79 & 25.13 \\
GPTQ      & 72.80 & 88.35 & 47.76 & 27.90 & 72.40 & 78.26 & 63.02 & 78.28 & 66.10 \\
GPTAQ     & 72.76 & 88.09 & 49.76 & 43.96 & 73.80 & 80.41 & 64.21 & 79.23 & 69.03 \\
\rowcolor{bgblue!10}\method   & 72.76 & 88.90 & 50.17 & 47.25 & 73.50 & 80.05 & 65.52 & 79.41 & \textbf{69.70}$_\textbf{(+0.67\%)}$ \\
\midrule

\multicolumn{10}{c}{\textit{Qwen2.5-VL-7B-Instruct-INT3}} \\
\midrule
Full Prec & 83.20 & 94.72 & 58.55 & 52.98 & 84.40 & 88.26 & 70.62 & 83.05 & 76.97 \\
GPTQ      & 62.44 & 89.33 & 42.31 & 27.33 & 77.70 & 79.39 & 66.58 & 78.38 & 65.43 \\
GPTAQ     & 63.68 & 90.41 & 47.18 & 36.49 & 79.60 & 80.52 & 66.40 & 77.94 & 67.78 \\
\rowcolor{bgblue!10}\method   & 65.32 & 91.36 & 45.50 & 37.89 & 80.00 & 81.70 & 67.32 & 79.28 & \textbf{68.55}$_\textbf{(+0.77\%)}$ \\
\midrule

\multicolumn{10}{c}{\textit{Qwen2.5-VL-32B-Instruct-INT3}} \\
\midrule
Full Prec & 69.24 & 92.51 & 60.21 & 60.37 & 80.30 & 92.69 & 71.67 & 77.06 & 75.51 \\
GPTQ  & 43.88 & 88.26 & 51.77 & 47.76 & 76.50 & 87.22 & 64.87 & 73.88 & 74.20 \\
GPTAQ & 52.52 & 88.72 & 52.13 & 42.67 & 77.40 & 79.46 & 67.89 & 72.88 & 73.68 \\
\rowcolor{bgblue!10}\method & 49.84 & 87.47 & 53.05 & 43.45 & 76.90 & 83.92 & 65.96 & 73.44 & \textbf{74.31}$_\textbf{(+0.11\%)}$ \\
\midrule

\multicolumn{10}{c}{\textit{LLaVA-OneVision-7B-INT3}} \\
\midrule
Full Prec & 80.08 & 87.09 & 57.39 & 53.93 & 62.10 & 89.98 & 64.82 & 75.96 & 71.42 \\
GPTQ      & 74.24 & 77.82 & 46.47 & 42.59 & 56.60 & 85.26 & 60.65 & 71.90 & 64.44 \\
GPTAQ     & 75.72 & 78.07 & 46.48 & 42.94 & 58.30 & 85.97 & 61.84 & 72.69 & 65.25 \\
\rowcolor{bgblue!10}\method & 74.56 & 78.21 & 49.12 & 45.94 & 59.20 & 86.37 & 61.05 & 73.66 & \textbf{66.01}$_\textbf{(+0.76\%)}$ \\

\bottomrule
\end{tabular}

%% file: tables/int3g128_exp.tex
\begin{tabular}{>{
    \raggedright\arraybackslash}m{2.0cm}
    *{8}{>{\centering\arraybackslash}m{1.2cm}}
    >{\raggedright\arraybackslash}m{1.5cm}
}
\toprule
\textbf{Method} & \textbf{Chart} & \textbf{Doc}$^\text{val}$ & \textbf{MME}$_\text{en}$ & \textbf{MME}$_\text{cn}$ & \textbf{OCR} & \textbf{SciQA} & \textbf{Seed}$^\text{2+}$ & \textbf{Text}$^\text{val}$ & \textbf{Avg} $(\uparrow)$\\
\midrule

\multicolumn{10}{c}{\textit{Qwen2-VL-7B-Instruct-INT3g128}} \\
\midrule
Full Prec & 81.44 & 93.89 & 57.30 & 56.09 & 80.90 & 85.55 & 69.26 & 82.02 & 75.81 \\
AWQ & 77.60 & 93.46 & 56.72 & 49.97 & 77.40 & 82.39 & 66.84 & 78.91 & 72.91 \\
MBQ & 77.20 & 93.68 & 56.79 & 50.50 & 76.60 & 82.24 & 65.74 & 78.67 & 72.68 \\
GPTQ & 79.08 & 92.21 & 54.47 & 49.21 & 79.80 & 83.16 & 66.80 & 81.02 & 73.22 \\
GPTAQ & 78.92 & 92.29 & 55.83 & 52.12 & 79.00 & 83.05 & 67.28 & 80.98 & 73.68\\
\rowcolor{bgblue!10}\method & 79.04 & 92.45 & 57.34 & 55.47 & 79.70 & 82.32 & 67.68 & 81.21 & \textbf{74.40}$_\textbf{(+0.72\%)}$ \\
\midrule

\multicolumn{10}{c}{\textit{Qwen2.5-VL-7B-Instruct-INT3g128}} \\
\midrule
Full Prec & 83.20 & 94.72 & 58.55 & 52.98 & 84.40 & 88.26 & 70.62 & 83.05 & 76.97 \\
GPTQ & 77.32 & 93.91 & 58.24 & 47.39 & 82.10 & 86.23 & 71.15 & 81.79 & 74.77 \\
GPTAQ & 75.36 & 93.76 & 56.32 & 44.33 & 83.40 & 86.28 & 69.74 & 82.25 & 73.93 \\
\rowcolor{bgblue!10}\method & 78.76 & 93.83 & 58.32 & 48.74 & 82.60 & 85.52 & 69.43 & 81.82 & \textbf{74.88}$_{\textbf{(+0.11\%)}}$  \\
\midrule

\multicolumn{10}{c}{\textit{LLaVA-OneVision-7B-INT3g128}} \\
\midrule
Full Prec & 80.08 & 87.09 & 57.39 & 53.93 & 62.10 & 89.98 & 64.82 & 75.96 & 71.42 \\
GPTQ & 78.40 & 84.53 & 55.20 & 49.64 & 60.30 & 88.85 & 63.46 & 74.55 & 69.37 \\
GPTAQ & 78.64 & 84.06 & 54.82 & 50.53 & 59.80 & 88.26 & 64.82 & 74.76 & 69.46 \\
\rowcolor{bgblue!10}\method & 77.92 & 83.89 & 54.26 & 51.48 & 62.10 & 88.42 & 63.99 & 74.74 & \textbf{69.60}$_\textbf{(+0.14\%)}$   \\

\bottomrule
\end{tabular}

%% file: tables/int2g128_exp.tex
\begin{tabular}{>{
    \raggedright\arraybackslash}m{2.0cm}
    *{8}{>{\centering\arraybackslash}m{1.2cm}}
    >{\raggedright\arraybackslash}m{1.5cm}
}
\toprule
\textbf{Method} & \textbf{Chart} & \textbf{Doc}$^\text{val}$ & \textbf{MME}$_\text{en}$ & \textbf{MME}$_\text{cn}$ & \textbf{OCR} & \textbf{SciQA} & \textbf{Seed}$^\text{2+}$ & \textbf{Text}$^\text{val}$ & \textbf{Avg} $(\uparrow)$\\
\midrule

\multicolumn{10}{c}{\textit{Qwen2-VL-7B-Instruct-INT2g128}} \\
\midrule
AWQ & 0.00 & 0.13 & 11.58 & 7.47 & 0.60 & 2.26 & 9.18 & 0.11 & 3.92 \\
MBQ & 0.00 & 0.06 & 12.46 & 7.94 & 0.90 & 3.14 & 9.00 & 0.06 & 4.19 \\
GPTQ & 56.44 & 74.90 & 41.33 & 34.11 & 62.60 & 50.37 & 51.78 & 71.93 & 55.43 \\
GPTAQ & 56.08 & 72.57 & 37.91 & 33.80 & 61.30 & 59.70 & 51.25 & 67.98 & 55.07 \\
\rowcolor{bgblue!10}\method & 55.32 & 75.76 & 41.97 & 35.59 & 62.50 & 62.32 & 53.80 & 74.80 & \textbf{57.76}$_\textbf{(+2.33\%)}$ \\
\midrule

\multicolumn{10}{c}{\textit{Qwen2.5-VL-7B-Instruct-INT2g128}} \\
\midrule
GPTQ & 57.72 & 78.79 & 44.94 & 13.89 & 70.80 & 55.77 & 48.40 & 74.36 & 55.58\\
GPTAQ & 53.48 & 74.82 & 39.38 & 3.60  & 67.70 & 2.95  & 17.83 & 73.79 & 41.69 \\
\rowcolor{bgblue!10}\method & 59.68 & 73.20 & 41.27 & 30.34 & 69.30 & 62.72 & 57.27 & 65.88 & \textbf{57.46}$_\textbf{(+1.88\%)}$ \\
\midrule

\multicolumn{10}{c}{\textit{LLaVA-OneVision-7B-INT2g128}} \\
\midrule
GPTQ & 61.08 & 62.62 & 40.29 & 31.35 & 48.60 & 67.88 & 51.25 & 67.29 & 53.80\\
GPTAQ & 62.12 & 62.81 & 42.59 & 32.45 & 49.90 & 66.12 & 50.94 & 67.35 & 54.29\\
\rowcolor{bgblue!10}\method   & 62.76 & 64.82 & 38.66 & 31.38 & 51.40 & 69.04 & 53.32 & 68.20 & \textbf{54.95}$_\textbf{(+0.66\%)}$\\

\bottomrule
\end{tabular}

%% file: tables/video_text_results.tex
\begin{tabular}{l cc ccc}
\toprule
\multirow{2}{*}{\textbf{Method}} & \multicolumn{2}{c}{\textbf{Text-heavy}} & \multicolumn{3}{c}{\textbf{Vision-heavy}} \\
\cmidrule(lr){2-3} \cmidrule(lr){4-6}
 & \textbf{HellaS} & \textbf{TCaps} & \textbf{MMMU} & \textbf{VMME} & \textbf{LVB} \\
\midrule
Full Prec & 77.63 & 152.00 & 50.44 & 57.78 & 56.92 \\
GPTQ      & 51.29 & 55.06  & 32.22 & 43.68 & 44.26 \\
GPTAQ     & 49.82 & 46.52  & 22.25 & 44.04 & 44.73 \\
\rowcolor{bgblue!10}SalQ-VLM & \textbf{53.19} & \textbf{55.90} & \textbf{33.67} & \textbf{45.70} & \textbf{46.81} \\
\bottomrule
\end{tabular}

%% file: tables/pruning_comparison.tex
\begin{tabular}{lccccc}
\toprule
\multirow{2}{*}{\textbf{Method}} & \multirow{2}{*}{\textbf{Full}} & \multicolumn{2}{c}{\textbf{FastV}~\citep{chen2024fastv}} & \multicolumn{2}{c}{\textbf{ToMe}~\citep{bolya2022tome}} \\
\cmidrule(lr){3-4} \cmidrule(lr){5-6}
 & & 50\% & 70\% & 50\% & 80\% \\
\midrule
FP & 50.67 & 49.44 & 49.00 & 49.78 & 49.56 \\
GPTQ & 32.78 & 32.33 & 32.67 & 32.44 & 31.67 \\
\rowcolor{bgblue!10}SalQ-VLM & \textbf{35.11} & \textbf{35.33} & \textbf{34.22} & \textbf{35.89} & \textbf{35.78} \\
\bottomrule
\end{tabular}

%% file: tables/ablation_importace_factor_type_compact.tex
\begin{tabular}{
    >{\centering\arraybackslash}m{3.0cm}
    >{\centering\arraybackslash}m{5.0cm}
    >{\centering\arraybackslash}m{1.6cm}
}
\toprule
\textbf{Factor Source} & \textbf{Style} & \textbf{Avg Acc} $(\uparrow)$\\
\midrule
Constant & $\operatorname{Diag}({\mathbf{1}})$ & 69.03 \\
Attention score & FastV~\citep{chen2024fastv} & 68.29 \\
Attention score & PACT~\citep{dhouib2025pact} & 67.07 \\
\rowcolor{bgblue!10}Gradient & \method (ours) & \textbf{69.70} \\
\bottomrule
\end{tabular}

%% file: tables/ablation_back_granularity_compact.tex
\begin{tabular}{
    >{\centering\arraybackslash}m{4.0cm}
    >{\centering\arraybackslash}m{1.5cm}
    >{\centering\arraybackslash}m{1.5cm}
    >{\centering\arraybackslash}m{1.4cm}
}
\toprule
\textbf{Loss Type} & \textbf{Avg Acc} $(\uparrow)$ & \textbf{GPU Hours} & \textbf{GPU Mem} \\
\midrule
$\gL_{\text{Network}}=\operatorname{CE}(\cdot)$ & 68.64\% & 0.70 & $\sim$30GB \\
$\gL_{\text{Layer}}=\operatorname{MSE}(\cdot)$ & 67.77\% & 0.29 & $\sim$300GB \\
\rowcolor{bgblue!10}$\gL_{\text{Block}}=\operatorname{MSE}(\cdot)$ & 69.70\% & 0.21 & $\sim$30GB \\ 
\bottomrule
\end{tabular}

%% file: tables/ablation_layer_selection_compact.tex
\begin{tabular}{
    >{\centering\arraybackslash}m{2.0cm}
    >{\centering\arraybackslash}m{3.0cm}
    >{\centering\arraybackslash}m{2.0cm}
}
\toprule
\textbf{Loss Term} & \textbf{Enhanced Layers} & \textbf{Avg Acc} $(\uparrow)$\\
\midrule
- & $\varnothing$ & 69.03\% \\
$||\Delta \text{Attn}(\rmX)||_2^2$ & $\{\texttt{x\_proj}\}_{\texttt{x=q/k/v}}$ & 55.54\% \\
$||\Delta \rmY||_2^2$ & $\{\texttt{x\_proj}\}_{\texttt{x=q/k/v}}$ & 56.21\% \\
\rowcolor{bgblue!10} $||\Delta \rmY||_2^2$ & $\{\texttt{x\_proj}\}_{\texttt{x=q/k/v/o}}$ & \textbf{69.70\%} \\
\bottomrule
\end{tabular}

%% file: tables/ablation_precursor.tex
\begin{tabular}{l cc cc}
\toprule
\multirow{2}{*}{\textbf{Precursor}} & \multicolumn{2}{c}{\textbf{Qwen2-VL-7B}} & \multicolumn{2}{c}{\textbf{Qwen2.5-VL-7B}} \\
\cmidrule(lr){2-3} \cmidrule(lr){4-5}
 & \textit{w/o} & \textit{w/} & \textit{w/o} & \textit{w/} \\
\midrule
\textcolor{gray}{AWQ}   & \textcolor{gray}{3.92}  & \textcolor{gray}{4.07}           & \textcolor{gray}{2.88}     & \textcolor{gray}{4.21}              \\
GPTQ  & 55.43 & \textbf{57.76} & 55.58 & \textbf{57.46} \\
GPTAQ & 55.07 & \textbf{56.51} & 41.69 & \textbf{52.10} \\
\bottomrule
\end{tabular}

%% file: tables/speedup.tex
\begin{tabular}{llccc}
\toprule
\textbf{Linear} & \textbf{$\mathbf{W}$ Size}  & \textbf{FP16} ($\mu s$) & \textbf{INT3} ($\mu s$) & \textbf{Speedup} \\
\midrule
\texttt{q/k/v\_proj}    & $3584 \times 4608$   & 38.97  & 4.17  & $9.34\times$ \\
\texttt{out\_proj}    & $3584 \times 3584$   & 39.00  & 4.17  & $9.34\times$ \\
\texttt{gate/up\_proj} & $3584 \times 37888$  & 175.17 & 18.46 & $9.49\times$ \\
\texttt{down\_proj}   & $18944 \times 3584$  & 96.39  & 10.47 & $9.20\times$ \\
\bottomrule
\end{tabular}